%% file: main.tex
\documentclass{article} 
\usepackage{iclr2026_conference,times}
\usepackage[T1]{fontenc}
\usepackage[utf8]{inputenc}

\input{math_commands.tex}

\usepackage{mathrsfs}
\usepackage{graphicx}
\usepackage{hyperref}
\usepackage{xurl}
\usepackage{placeins}
\hypersetup{hidelinks,
  pdftitle={Learning to Disprove: Formal Counterexample Generation with Large Language Models},
  pdfauthor={Zenan Li, Zhaoyu Li, Kaiyu Yang, Xiaoxing Ma, Zhendong Su}}
\usepackage{enumitem}
\usepackage{arydshln} 
\usepackage{threeparttable}
\usepackage{booktabs}
\usepackage{subcaption}
\usepackage{multirow}
\usepackage{xspace}
\usepackage[table]{xcolor}
\usepackage{tikz}
\usepackage{makecell}

\usepackage{inconsolata}

\usepackage{listings}
\usepackage[most]{tcolorbox}

\input{lstlean.tex}

\definecolor{keywordcolor}{rgb}{0.7, 0.1, 0.1}   
\definecolor{tacticcolor}{rgb}{0.0, 0.1, 0.6}    
\definecolor{commentcolor}{rgb}{0.4, 0.4, 0.4}   
\definecolor{symbolcolor}{rgb}{0.0, 0.1, 0.6}    
\definecolor{sortcolor}{rgb}{0.1, 0.5, 0.1}      
\definecolor{attributecolor}{rgb}{0.7, 0.1, 0.1} 

\lstdefinestyle{lean}{
  language=lean,
  frame=single,
  breakindent=0pt,
  basicstyle=\ttfamily,
}
\lstdefinestyle{leaninline}{
  language=lean,
  breakindent=0pt,
  frame=single,
  basicstyle=\ttfamily,
}
\newcommand{\leaninline}[1]{\lstinline[style=leaninline]{#1}}

\usepackage{algorithm,algpseudocode}

\title{Learning to Disprove:\\Formal Counterexample Generation\\with Large Language Models}

\author{Zenan Li\textsuperscript{1}\thanks{Correspondence to: \href{mailto:zenan.li@inf.ethz.ch}{\texttt{zenan.li@inf.ethz.ch}}} \quad Zhaoyu Li\textsuperscript{2} \quad Kaiyu Yang\textsuperscript{3} \quad Xiaoxing Ma\textsuperscript{4} \quad Zhendong Su\textsuperscript{1} \\
\textsuperscript{1}ETH Zurich \quad \textsuperscript{2}University of Toronto \\
\textsuperscript{3}MiroMind \quad \textsuperscript{4}Nanjing University \\
}

\iclrfinalcopy 
\begin{document}
\maketitle

\begin{abstract}
Mathematical reasoning demands two critical, complementary skills: constructing rigorous proofs for true statements and discovering counterexamples that disprove false ones.
However, current AI efforts in mathematics focus almost exclusively on proof construction, often neglecting the equally important task of finding counterexamples. 
In this paper, we address this gap by fine-tuning large language models (LLMs) to reason about and generate counterexamples.
We formalize this task as formal counterexample generation, which requires LLMs not only to propose candidate counterexamples but also to produce formal proofs that can be automatically verified in the Lean 4 theorem prover.
To enable effective learning, we introduce a symbolic mutation strategy that constructs candidate counterexample-generation tasks by systematically extracting theorems and removing selected hypotheses.
Together with curated datasets, this strategy enables a multi-reward expert iteration framework that substantially enhances both the effectiveness and efficiency of training LLMs for counterexample generation and theorem proving.
Experiments on three newly collected benchmarks validate the advantages of our approach, showing that the mutation strategy and training framework yield substantial performance gains.
\end{abstract}

\input{sections/001_introduction}
\input{sections/002_problem}

\input{sections/003_solution}

\input{sections/004_experiment}
\input{sections/005_relatedwork}

\section{Limitations}
While our proposed framework demonstrates strong potential in counterexample search and formal proof generation, several directions remain for further improvement and extension.

\textbf{Data quality of synthetic data.}
Compared with manually collecting and labeling problems and solutions, symbolic mutation can generate large volumes of data at extremely low cost.
However, data quality remains an extensive concern, as the generated data often lacks careful selection.
This results in the downgrade of training efficiency.
For instance, as shown in our training curves (Figure~\ref{fig:curves}), LLM fine-tuning converges within roughly half an epoch, suggesting that a substantial portion of the generated data is redundant.
Although prior studies have explored methods to improve synthetic data quality~\citep{muennighoff2025s1,xia2024less,wang2024greats}, effectively incorporating these techniques into formal counterexample generation remains an open challenge.

\textbf{Limited budget of model training.}
We provide additional case studies in Appendix~\ref{app:case_study}, which demonstrate that the limited capability (or size) of the LLM significantly compromises the performance of our framework.
First, the 8B informal reasoning model frequently fails to produce correct results, even for simple calculations. Employing a larger model or incorporating tool-use strategies could help alleviate this issue.
Second, the 7B formal reasoning model often struggles to follow instructions and fails to utilize the provided counterexamples when deriving formal proofs. This leads to inaccurate feedback rewards that undermine overall performance.
While these limitations could likely be mitigated by using larger LLMs, resource constraints prevent us from exploring this direction at present. We leave it for future work and further investigation.

\section{Conclusion}

In this work, we investigated the crucial yet largely underexplored problem of formal counterexample generation with reasoning large language models.
We identified and discussed two key challenges, data scarcity and sparse reward signals, that hinder the development of models specialized for this task.
To address these issues, we introduced an integrated framework that combines a symbolic mutation strategy for large-scale counterexample synthesis with a multi-reward scheme for more effective and efficient model training. 
Experiments on three newly constructed benchmarks demonstrate that our approach substantially outperforms existing state-of-the-art LLMs in counterexample generation.
We believe that our proposed framework opens up a promising direction for enhancing the self-reflective and self-corrective reasoning capabilities of LLMs, while also providing a practical copilot for validating mathematical conjectures.


\section*{Acknowledgments}
We thank Terry Ding for pointing out an error in our formulation of the counterexample target.
\section*{Ethics Statement}
This work focuses on formal counterexample generation and does not involve human subjects or sensitive data. The datasets employed are publicly available and widely used in prior research. We do not anticipate any ethical concerns or harmful applications arising from this study.

\section*{Reproducibility Statement}
We provide detailed descriptions of the training and inference setup in the main text (Section~\ref{sec:experiment}) and Appendix~\ref{app:setting}. All datasets and preprocessing steps are thoroughly documented. Source code is available at \url{https://doi.org/10.6084/m9.figshare.31741879} to ensure full reproducibility.

\FloatBarrier
\bibliography{references}
\bibliographystyle{iclr2026_conference}

\clearpage
\appendix

\input{sections/00x_appendix}

\end{document}

%% file: math_commands.tex
\usepackage{amsmath,amsfonts,bm}

\def\eqref#1{equation~\ref{#1}}

\def\1{\bm{1}}

\DeclareMathAlphabet{\mathsfit}{\encodingdefault}{\sfdefault}{m}{sl}
\SetMathAlphabet{\mathsfit}{bold}{\encodingdefault}{\sfdefault}{bx}{n}



%% file: lstlean.tex
\lstdefinelanguage{lean} {
mathescape=false,
texcl=false,
morekeywords=[1]{
import, prelude, protected, private, noncomputable, definition, meta, renaming,
hiding, parameter, parameters, begin, constant, constants,
lemma, variable, variables, theory,
print, theorem, example,
open, as, export, override, axiom, axioms, inductive, with,
structure, record, universe, universes,
alias, help, precedence, reserve, declare_trace, add_key_equivalence,
match, infix, infixl, infixr, notation, postfix, prefix, instance,
eval, reduce, check, end, this,
using, using_well_founded, namespace, section,
attribute, local, set_option, extends, include, omit, class,
raw, replacing,
calc, have, show, suffices, by, in, at, let, forall, Pi, fun,
exists, if, dif, then, else, assume, obtain, from, register_simp_ext, unless, break, continue,
mutual, do, def, run_cmd, const,
partial, mut, where, macro, syntax, deriving,
return, try, catch, for, macro_rules, declare_syntax_cat, abbrev, sorry},
morekeywords=[2]{Sort, Type, Prop},
morekeywords=[3]{
assumption,
apply, intro, intros, allGoals,
generalize, clear, revert, done, exact,
refine, repeat, cases, rewrite, rw,
simp, simp_all, contradiction,
constructor, injection,
induction, 
prove_with, linarith, nlinarith, norm_num, contrapose, rfl, by_contra, push_neg, gcongr, aesop, field_simp, rfl, norm_cast, ring, ring_nf, positivity, omega, subst, rename_i, rcases, convert, ext, tauto, native_decide, use, by_cases,
},
literate=
{@@u03b1@@}{{\ensuremath{\mathrm{\alpha}}}}1
{@@u03b2@@}{{\ensuremath{\mathrm{\beta}}}}1
{@@u03b3@@}{{\ensuremath{\mathrm{\gamma}}}}1
{@@u03b4@@}{{\ensuremath{\mathrm{\delta}}}}1
{@@u03b5@@}{{\ensuremath{\mathrm{\varepsilon}}}}1
{@@u03b6@@}{{\ensuremath{\mathrm{\zeta}}}}1
{@@u03b7@@}{{\ensuremath{\mathrm{\eta}}}}1
{@@u03b8@@}{{\ensuremath{\mathrm{\theta}}}}1
{@@u03b9@@}{{\ensuremath{\mathrm{\iota}}}}1
{@@u03ba@@}{{\ensuremath{\mathrm{\kappa}}}}1
{@@u03bc@@}{{\ensuremath{\mathrm{\mu}}}}1
{@@u03bd@@}{{\ensuremath{\mathrm{\nu}}}}1
{@@u03be@@}{{\ensuremath{\mathrm{\xi}}}}1
{@@u03c0@@}{{\ensuremath{\mathrm{\mathnormal{\pi}}}}}1
{@@u03c1@@}{{\ensuremath{\mathrm{\rho}}}}1
{@@u03c3@@}{{\ensuremath{\mathrm{\sigma}}}}1
{@@u03c4@@}{{\ensuremath{\mathrm{\tau}}}}1
{@@u03c6@@}{{\ensuremath{\mathrm{\varphi}}}}1
{@@u03c7@@}{{\ensuremath{\mathrm{\chi}}}}1
{@@u03c8@@}{{\ensuremath{\mathrm{\psi}}}}1
{@@u03c9@@}{{\ensuremath{\mathrm{\omega}}}}1
{@@u0393@@}{{\ensuremath{\mathrm{\Gamma}}}}1
{@@u0394@@}{{\ensuremath{\mathrm{\Delta}}}}1
{@@u0398@@}{{\ensuremath{\mathrm{\Theta}}}}1
{@@u039b@@}{{\ensuremath{\mathrm{\Lambda}}}}1
{@@u03a3@@}{{\ensuremath{\mathrm{\Sigma}}}}1
{@@u03a6@@}{{\ensuremath{\mathrm{\Phi}}}}1
{@@u039e@@}{{\ensuremath{\mathrm{\Xi}}}}1
{@@u03a8@@}{{\ensuremath{\mathrm{\Psi}}}}1
{@@u03a9@@}{{\ensuremath{\mathrm{\Omega}}}}1
{@@u2135@@}{{\ensuremath{\aleph}}}1
{@@u2264@@}{{\ensuremath{\leq}}}1
{@@u2265@@}{{\ensuremath{\geq}}}1
{@@u2260@@}{{\ensuremath{\neq}}}1
{@@u2248@@}{{\ensuremath{\approx}}}1
{@@u2261@@}{{\ensuremath{\equiv}}}1
{@@u2243@@}{{\ensuremath{\simeq}}}1
{@@u2264@@}{{\ensuremath{\leq}}}1
{@@u2265@@}{{\ensuremath{\geq}}}1
{@@u2202@@}{{\ensuremath{\partial}}}1
{@@u2206@@}{{\ensuremath{\triangle}}}1 
{@@u222b@@}{{\ensuremath{\int}}}1
{@@u2211@@}{{\ensuremath{\mathrm{\Sigma}}}}1
{@@u03a0@@}{{\ensuremath{\mathrm{\Pi}}}}1
{@@u22a5@@}{{\ensuremath{\perp}}}1
{@@u221e@@}{{\ensuremath{\infty}}}1
{@@u2202@@}{{\ensuremath{\partial}}}1
{@@u2213@@}{{\ensuremath{\mp}}}1
{@@u00b1@@}{{\ensuremath{\pm}}}1
{@@u00d7@@}{{\ensuremath{\times}}}1
{@@u2295@@}{{\ensuremath{\oplus}}}1
{@@u2297@@}{{\ensuremath{\otimes}}}1
{@@u229e@@}{{\ensuremath{\boxplus}}}1
{@@u2207@@}{{\ensuremath{\nabla}}}1
{@@u221a@@}{{\ensuremath{\sqrt}}}1
{@@u2b1d@@}{{\ensuremath{\cdot}}}1
{@@u2022@@}{{\ensuremath{\cdot}}}1
{@@u2218@@}{{\ensuremath{\circ}}}1
{@@u207b@@}{{\ensuremath{^{-}}}}1
{@@u25b8@@}{{\ensuremath{\blacktriangleright}}}1
{@@u2227@@}{{\ensuremath{\wedge}}}1
{@@u2228@@}{{\ensuremath{\vee}}}1
{@@u00ac@@}{{\ensuremath{\neg}}}1
{@@u22a2@@}{{\ensuremath{\vdash}}}1
{@@u27e8@@}{{\ensuremath{\langle}}}1
{@@u27e9@@}{{\ensuremath{\rangle}}}1
{@@u21a6@@}{{\ensuremath{\mapsto}}}1
{@@u2190@@}{{\ensuremath{\leftarrow}}}1
{<-}{{\ensuremath{\leftarrow}}}1
{@@u2192@@}{{\ensuremath{\rightarrow}}}1
{@@u2194@@}{{\ensuremath{\leftrightarrow}}}1
{@@u21d2@@}{{\ensuremath{\Rightarrow}}}1
{@@u27f9@@}{{\ensuremath{\Longrightarrow}}}1
{@@u21d0@@}{{\ensuremath{\Leftarrow}}}1
{@@u27f8@@}{{\ensuremath{\Longleftarrow}}}1
{@@u2229@@}{{\ensuremath{\cap}}}1
{@@u222a@@}{{\ensuremath{\cup}}}1
{@@u2282@@}{{\ensuremath{\subseteq}}}1
{@@u2286@@}{{\ensuremath{\subseteq}}}1
{@@u2284@@}{{\ensuremath{\nsubseteq}}}1
{@@u2288@@}{{\ensuremath{\nsubseteq}}}1
{@@u2283@@}{{\ensuremath{\supseteq}}}1
{@@u2287@@}{{\ensuremath{\supseteq}}}1
{@@u2285@@}{{\ensuremath{\nsupseteq}}}1
{@@u2289@@}{{\ensuremath{\nsupseteq}}}1
{@@u2208@@}{{\ensuremath{\in}}}1
{@@u2209@@}{{\ensuremath{\notin}}}1
{@@u220b@@}{{\ensuremath{\ni}}}1
{@@u220c@@}{{\ensuremath{\notni}}}1
{@@u2205@@}{{\ensuremath{\emptyset}}}1
{@@u2216@@}{{\ensuremath{\setminus}}}1
{@@u2020@@}{{\ensuremath{\dag}}}1
{@@u2115@@}{{\ensuremath{\mathbb{N}}}}1
{@@u2124@@}{{\ensuremath{\mathbb{Z}}}}1
{@@u211d@@}{{\ensuremath{\mathbb{R}}}}1
{@@u211a@@}{{\ensuremath{\mathbb{Q}}}}1
{@@u2102@@}{{\ensuremath{\mathbb{C}}}}1
{@@u231e@@}{{\ensuremath{\llcorner}}}1
{@@u231f@@}{{\ensuremath{\lrcorner}}}1
{@@u2983@@}{{\ensuremath{\{\!|}}}1
{@@u2984@@}{{\ensuremath{|\!\}}}}1
{@@u2016@@}{{\ensuremath{\|}}}1
{@@u2081@@}{{\ensuremath{_1}}}1
{@@u2082@@}{{\ensuremath{_2}}}1
{@@u2083@@}{{\ensuremath{_3}}}1
{@@u2084@@}{{\ensuremath{_4}}}1
{@@u2085@@}{{\ensuremath{_5}}}1
{@@u2086@@}{{\ensuremath{_6}}}1
{@@u2087@@}{{\ensuremath{_7}}}1
{@@u2088@@}{{\ensuremath{_8}}}1
{@@u2089@@}{{\ensuremath{_9}}}1
{@@u2080@@}{{\ensuremath{_0}}}1
{@@u1d62@@}{{\ensuremath{_i}}}1
{@@u2c7c@@}{{\ensuremath{_j}}}1
{@@u2090@@}{{\ensuremath{_a}}}1
{@@u00b9@@}{{\ensuremath{^1}}}1
{@@u2099@@}{{\ensuremath{_n}}}1
{@@u2098@@}{{\ensuremath{_m}}}1
{@@u209a@@}{{\ensuremath{_p}}}1
{@@u2191@@}{{\ensuremath{\uparrow}}}1
{@@u2193@@}{{\ensuremath{\downarrow}}}1
{...}{{\ensuremath{\ldots}}}1
{@@u00b7@@}{{\ensuremath{\cdot}}}1
{@@u25b8@@}{{\ensuremath{\triangleright}}}1
{@@u2223@@}{{\ensuremath{\mid}}}1
{@@u00b9@@}{{\ensuremath{^1}}}1
{@@u00b2@@}{{\ensuremath{^2}}}1
{@@u00b3@@}{{\ensuremath{^3}}}1
{@@u2074@@}{{\ensuremath{^4}}}1
{@@u2075@@}{{\ensuremath{^5}}}1
{@@u2076@@}{{\ensuremath{^6}}}1
{@@u2077@@}{{\ensuremath{^7}}}1
{@@u2078@@}{{\ensuremath{^8}}}1
{@@u2079@@}{{\ensuremath{^9}}}1
{@@u2070@@}{{\ensuremath{^0}}}1
{@@u03a3@@}{{\color{symbolcolor}\ensuremath{\Sigma}}}1
{@@u03a0@@}{{\color{symbolcolor}\ensuremath{\Pi}}}1
{@@u2200@@}{{\color{symbolcolor}\ensuremath{\forall}}}1
{@@u2203@@}{{\color{symbolcolor}\ensuremath{\exists}}}1
{@@u03bb@@}{{\color{symbolcolor}\ensuremath{\mathrm{\lambda}}}}1
{\$}{{\color{symbolcolor}\$}}1
{:=}{{\color{symbolcolor}:=}}1
{=}{{\color{symbolcolor}=}}1
{<|>}{{\color{symbolcolor}<|>}}1
{<\$>}{{\color{symbolcolor}<\$>}}1
{+}{{\color{symbolcolor}+}}1
{*}{{\color{symbolcolor}*}}1,
morecomment=[s][\color{commentcolor}]{/-}{-/},
morecomment=[l][\itshape \color{commentcolor}]{--},
showstringspaces=false,
keepspaces=true,
morestring=[b]",
tabsize=3,
extendedchars=false,
sensitive=true,
breaklines=true,
breakatwhitespace=true,
basicstyle=\ttfamily\small,
captionpos=b,
columns=[l]fullflexible,
identifierstyle={\ttfamily\color{black}},
keywordstyle=[1]{\ttfamily\color{keywordcolor}},
keywordstyle=[2]{\ttfamily\color{sortcolor}},
keywordstyle=[3]{\ttfamily\color{tacticcolor}},
keywordstyle=[4]{\ttfamily\color{attributecolor}},
stringstyle=\ttfamily,
commentstyle={\ttfamily\footnotesize },
}

%% file: sections/001_introduction.tex
\section{Introduction}

Endowing machines with mathematical reasoning capabilities has long represented one of the most fundamental and challenging frontiers in artificial intelligence research~\citep{ahn2024large}. 
In recent years, with the emergence of reasoning large language models (LLMs) such as OpenAI-o1~\citep{openai2023o1preview}, DeepSeek-R1~\citep{guo2025deepseek}, and Gemini-2.5-Pro~\citep{comanici2025gemini}, significant advances have been achieved in this area. 
Furthermore, several formal reasoning LLMs, including Seed prover~\citep{chen2025seed}, Kimina prover~\citep{wang2025kimina}, and Goedel prover~\citep{lin2025goedel}, have also demonstrated strong performance in generating formal proofs and interacting with theorem provers~\citep{de2015lean, barras1999coq, paulson1993isabelle}. 
The impressive capabilities exhibited by these models open up new avenues for addressing advanced and complex mathematical problems.
However, existing studies often emphasize the logical deduction of LLMs; their capability to find counterexamples remains underexplored~\citep{li2025one}. 

In mathematics, counterexamples are far more than mere exceptions~\citep{Barahmand2019On}; they play a vital role in theory development~\citep{lander1967survey}, conjecture refinement~\citep{haselgrove1958disproof}, and educational enhancement~\citep{zazkis2008makes}. 
For instance, since a conjecture must hold for all possible inputs, identifying anomalous examples and special cases is a crucial step for formulating valuable conjectures.
Moreover, for LLM reasoning, its effectiveness largely depends on how well models can reflect upon and evaluate their intermediate reasoning steps~\citep{renze2024self}.
Hence, counterexample generation becomes equally vital: it equips models with the ability to self-verify their logical processes, thereby strengthening both their reasoning reliability and explanatory power.

In this paper, we focus on training LLMs to generate formal counterexamples.
Although casting the task into a formal setting requires the LLM to additionally provide a formal proof for each counterexample, 
it allows for the automatic evaluation of the proposed counterexample via the Lean 4 theorem prover.
Two significant challenges arise in this task:
First, there is a severe scarcity of training data. 
Currently, CounterMath~\citep{li2025one} stands as the sole dataset specifically curated for benchmarking counterexample generation, 
comprising only 1,216 natural language problems. 
This limited size renders it inadequate for practical LLM training.
Second, the reward signals during training are exceptionally sparse. 
When LLMs fail to produce correct counterexamples for complex problems, the training reward vanishes, hindering further improvement in model performance and resulting in a plateau at a low success rate.

\begin{figure}[t]
    \centering
    \includegraphics[width=1.0\textwidth]{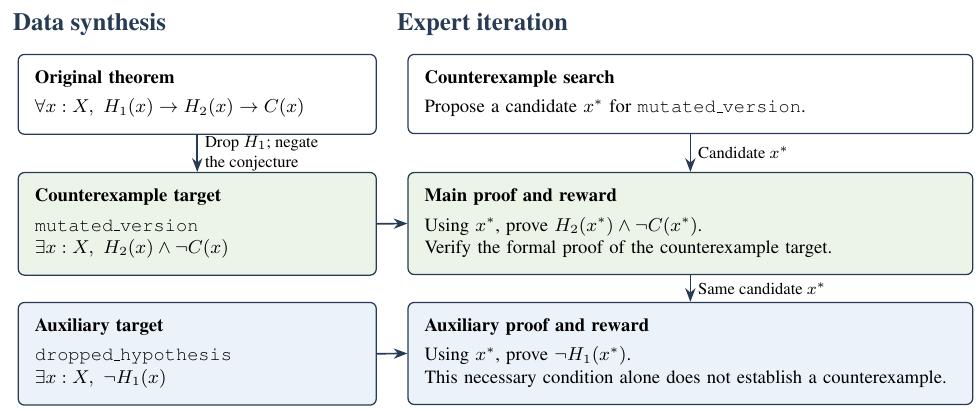}  
    \caption{
    Framework of counterexample training. 
    In the data synthesis stage, the symbolic mutation drops the hypothesis of a provable theorem, creating candidate counterexample problems.
    In the subsequent expert stage, two rewards are introduced based on whether proofs establish the counterexample target and the negation of the dropped hypothesis, boosting training effectiveness and efficiency. }

    \label{fig:framework}
\end{figure}

To address these challenges, we propose an integrated framework illustrated in Figure~\ref{fig:framework}. 
We design a new mutation strategy that drops selected hypotheses from provable theorems to construct candidate counterexample problems.
Through this mutation strategy, we synthesize a large problem dataset for counterexample search.
Furthermore, we introduce a multi-reward function based on verified proofs of the counterexample target and the negation of the discarded hypothesis,
which can provide auxiliary feedback even when the main counterexample target remains unsolved.

Experiments on three newly collected benchmarks demonstrate the effectiveness of our framework. 
Leveraging the proposed mutation strategy, we synthesized 575K candidate counterexample problems for fine-tuning.
Together with a curated large-scale dataset and a multi-reward training scheme, this allows us to train a model specialized for counterexample generation. 
We further establish three applicable tasks: counterexample search, verification of autoformalized results, and verification of reasoning steps.
Experimental results show that our fine-tuned model achieves substantial improvements on these tasks over current LLMs.
Notably, our fine-tuned model achieves a relative improvement of 47\% to 74\% in pass@1 success rate compared to the strongest baseline.


%% file: sections/002_problem.tex
\section{Formal Counterexample Generation}

The counterexample problem arises in disproving a universal mathematical conjecture of the form $\forall x, \mathtt{P}(x)$.~
By identifying a counterexample $x^*$, the conjecture can be invalidated through its negation $\exists x, \neg \mathtt{P}(x)$, which is established by proving the instantiated negation $\neg \mathtt{P}(x^*)$.
A classic example is the conjecture ``all continuous functions are differentiable.'' 
To refute this claim, one must demonstrate the existence of a function that is continuous but non-differentiable, e.g., the absolute value function $f(x)=|x|$.
Recently, automated mathematical research and knowledge discovery have garnered increasing attention~\citep{boiko2023autonomous, romera2024mathematical, novikov2025alphaevolve}.
As a result, the counterexample generation, as a specific yet crucial step in judging the correctness and completeness of newly proposed statements, has also evolved from being merely an incremental improvement to becoming pivotal in shaping the future of mathematical inquiry. 

There are several symbolic tools designed for finding counterexamples to conjectures, such as \emph{nitpick} in Isabelle~\citep{blanchette2010nitpick} and \emph{plausible} in Lean 4~\citep{lean2024plausible}. 
They often rely on sophisticated software engineering techniques like random testing or SAT/SMT solving~\citep{berard2013systems, ammann2017introduction}. 
Nonetheless, mathematics is fundamentally different from program proving, and the inherent complexity of higher-order logic hinders the effectiveness of these tools in both formulating problems and generating solutions.
A promising alternative lies in LLM-based counterexample generation. 
LLMs offer substantial expressiveness and strong reasoning capabilities, albeit with some sacrifice in soundness compared to symbolic tools.

In addition to using LLM-based counterexample generation as a powerful tool for examining mathematical conjectures, training LLMs to generate counterexamples offers a new perspective for boosting the LLM's reasoning capabilities.
Unlike traditional mathematical proofs that are typically derived through logical deduction, aligning well with existing chain-of-thought models~\citep{wei2022chain, chen2025towards}, counterexample generation involves a guess-and-check paradigm~\citep{young2009learning, redish2016vicarious}; the LLM first proposes a potential counterexample and then verifies its validity. 

\begin{figure}[t]
    \centering
    \includegraphics[width=0.95\textwidth]{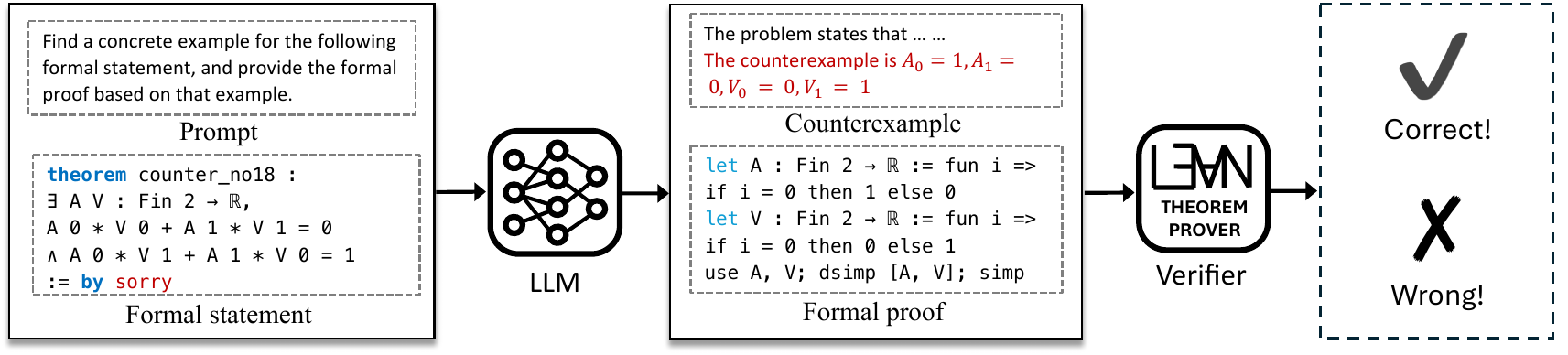}  
    \caption{Task of formal counterexample generation. This task requires the LLM first to perform informal reasoning to identify a valid counterexample for the given problem, and then generate the corresponding formal proof, which is automatically verified by theorem provers (e.g., Lean 4).}

    \label{fig:problem}
\end{figure}

Formal counterexample generation aims to establish a two-stage, informal-to-formal process.~
As illustrated in Figure~\ref{fig:problem}, given a formal problem, the LLM is first prompted to find a counterexample through informal reasoning.~
It then produces a formal proof for the theorem based on this derived counterexample.  
Finally, the validity of the proposed counterexample is confirmed if its corresponding formal proof is successfully verified by the Lean 4 theorem prover~\citep{de2015lean}.

This approach differs from other informal-to-formal reasoning methods, which mainly focus on creating an informal sketch followed by its formalization and completion~\citep{jiang2023draft, ren2025deepseek, cao2025reviving}.
In those methods, informal and formal reasoning are consistent and congruous with one another. 
In contrast, formal counterexample generation necessitates a sequential pattern of informal-to-formal reasoning. 
Specifically, the initial natural language reasoning serves as a counterexample proposal, transforming the abstract challenge of proving an existential statement into the more concrete task of formally verifying a specific instance.

Nevertheless, both stages of formal counterexample generation pose distinct challenges for current LLMs. 
The ``guess'' phase requires models to propose effective counterexamples by analyzing various edge scenarios, including corner cases and limit values. 
This particular capability remains an undeveloped aspect of their training~\citep{li2025one}. 
In the ``check'' phase, LLMs generate formal proofs, an endeavor complicated by the scarcity of training data for existential theorems~\citep{ying2024lean}, which significantly hinders the models' performance. 
By addressing these challenges, counterexample generation can become a powerful tool to enhance the reasoning capabilities of LLMs and assist mathematicians in verifying conjectures.



%% file: sections/003_solution.tex
\section{The Proposed Framework}

Our proposed approach consists of two stages: (1) counterexample problem synthesis and (2) multi-reward guided training. 
The first stage is devoted to creating a large-sized unlabeled dataset, i.e., the counterexample problem without a ground-truth answer and a formal proof. 
In the second stage, we carry out the expert iteration, where the LLM repeatedly generates counterexamples and formal proofs and then is fine-tuned on the successful ones. 
We will outline each stage as follows.

\subsection{Counterexample Problem Synthesis}

We first collect a substantial number of formal theorems to serve as foundational seeds. 
These seed theorems must meet two criteria: they should be (1) formally provable and (2) expressed in a universal format (e.g., $\forall x$).
In addition to sourcing these from existing Lean 4 formal libraries, like Mathlib, we extract intermediate subgoals or lemmas from formal proofs generated by LLMs. 
Specifically, given a formal proof, we decompose it into a series of proof steps, and categorize each proof step by its proving style.
If a proof step is stated in declarative style (i.e., using \emph{have} or \emph{suffices} tactic), we directly transform it into a new theorem.
Conversely, for a proof step stated in a procedural style (using tactics such as \emph{simp} and \emph{rw}), we analyze and compare the proof states before and after applying such a tactic to establish a new theorem.

Next, we design a mutation-based data synthesis method. 
To elaborate, let us start with a formally provable, universally stated theorem, which can be represented as
\begin{tcolorbox}[breakable, enhanced jigsaw, top=5pt, bottom=5pt, boxrule=0.5pt]
\begin{lstlisting}[style=lean, frame=none]
theorem original_version (x : X) : H@@u2081@@ x @@u2192@@ H@@u2082@@ x @@u2192@@ C x 
\end{lstlisting}
\end{tcolorbox}
Here, $H_1$ and $H_2$ are two hypotheses, and  $C$ is the conclusion. 
This theorem asserts that for any element $x$ in the space $X$, if both $H_1(x)$ and $H_2(x)$ hold true, then $C(x)$ follows.

Discarding one of the hypotheses (for example, $H_1(x)$) yields the universally quantified conjecture $\forall x : X,\ H_2(x) \to C(x)$.
If $H_1$ is essential to the validity of the original theorem, this conjecture is false. Its negation is classically equivalent to $\exists x : X,\ H_2(x) \land \neg C(x)$: a counterexample must satisfy the retained hypothesis and falsify the conclusion.
We formalize this counterexample target as
\begin{tcolorbox}[breakable, enhanced jigsaw, top=5pt, bottom=5pt, boxrule=0.5pt]
\begin{lstlisting}[style=lean, frame=none]
theorem mutated_version : @@u2203@@ x : X, H@@u2082@@ x @@u2227@@ @@u00ac@@ C x
\end{lstlisting}
\end{tcolorbox}

Before mutation, we use Lean 4 to identify and remove hypotheses unused in the original proof~\citep{lean_unused_variables}. This is a heuristic filter: a hypothesis used in a particular proof need not be logically necessary for the theorem.
Mutation therefore does not guarantee that the resulting conjecture is false or that a counterexample exists. A generated target is confirmed satisfiable only when a witness is verified.
Compared with LLM-based problem augmentation~\citep{wang2024survey, yu2023metamath, huang2024mustard}, this symbolic strategy provides a scalable way to generate candidate counterexample problems for training.

\subsection{Multi-reward Guided Training}

We follow the standard expert iteration pipeline~\citep{anthony2017thinking} to train the LLMs with unlabeled data. 
In this framework, the model is repeatedly applied to perform large-scale inference on counterexample problems, from which candidate solutions and formal proofs are collected. The correct ones are then used to further refine the model through supervised fine-tuning.
A key limitation, noted in prior work, is that LLMs often fail to produce correct results on difficult problems~\citep{polu2022formal, wu2024internlm2, dong2025stp}, restricting training to a cycle of generating and fine-tuning only on simpler examples.
This issue is essentially the sparse reward problem in classic reinforcement learning~\citep{hare2019dealing, riedmiller2018learning}, commonly alleviated through techniques such as reward shaping~\citep{laud2004theory} or curriculum learning~\citep{bengio2009curriculum}.

Rather than adopting these conventional approaches, we propose using a multi-reward strategy to effectively address this limitation.
A witness $x^*$ for the counterexample target $\exists x, H_2(x) \land \neg C(x)$ must also satisfy $\neg H_1(x^*)$: otherwise, the original theorem and $H_2(x^*)$ would imply $C(x^*)$, a contradiction.
This leads to the formation of a new theorem
\begin{tcolorbox}[breakable, enhanced jigsaw, top=5pt, bottom=5pt, boxrule=0.5pt]
\begin{lstlisting}[style=lean, frame=none]
theorem dropped_hypothesis : @@u2203@@ x : X, @@u00ac@@ H@@u2081@@ x
\end{lstlisting}
\end{tcolorbox}

To score a candidate $x^*$, the LLM is prompted to prove \emph{dropped\_hypothesis} and \emph{mutated\_version} using that candidate as the witness.
These proofs are subsequently verified by the theorem prover, which yields two distinct rewards for the counterexample.
When the generated proofs use the proposed witness, the auxiliary reward corresponds to proving $\neg H_1(x^*)$, while the main reward corresponds to proving $H_2(x^*) \land \neg C(x^*)$. The auxiliary condition alone does not establish a counterexample. Although the prompts request the same proposed witness, verifying a proof of an existential target alone does not guarantee that the proof uses that witness. This is a limitation of the feedback mechanism, discussed further under limited model capability in the Limitations section.
The auxiliary target can provide a training signal when its proof succeeds even if the main target remains unsolved; it does not guarantee a nonzero reward or a valid counterexample.
Ultimately, we combine these two rewards as the final sample weight for the supervised fine-tuning.

Additionally, we include the training of LLMs for generating formal proofs of counterexamples in the expert iteration. 
Besides collecting the formal proof of \emph{mutated\_version}, the formal proof of \emph{dropped\_hypothesis} can also be used for fine-tuning. 
However, since formally proving the necessary condition is often easier, the model might fall into a low-difficulty trap, frequently generating shorter proofs. 
To mitigate this, the sample weight of the auxiliary proof should be reduced, and other techniques, e.g., group relative policy optimization (GRPO)~\citep{shao2024deepseekmath}, can also be employed to encourage the generation of longer formal proofs.

\subsection{Overall Framework}

Formally, let us define the seed dataset as $\{\mathcal{T}_i\}_{i=1}^N$, where $\mathcal{T}_i$ stands for a seed theorem. 
For every theorem $\mathcal{T}_i$, mutation yields the counterexample target $\mathcal{M}_i$ and the auxiliary target $\mathcal{H}_i$ expressing the negation of the dropped hypothesis.
We introduce two LLMs, represented as $q_{\phi}$ and $q_{\psi}$, for generating counterexamples and formal proofs, respectively. 
For a given counterexample problem $\mathcal{M}_i$, we first prompt the LLM to propose a counterexample $x_i \sim q_{\phi}(x_i \mid \mathcal{M}_i)$.
Next, the formal proof $\pi_i^{\mathcal{M}}$ is constructed by integrating the problem and the proposed counterexample, i.e., $\pi_i^{\mathcal{M}} \sim q_{\psi}(\pi_i^{\mathcal{M}} \mid \mathcal{M}_i, x_i)$.
In addition, the LLM provides the formal proof $\pi_i^{\mathcal{H}}$ for the auxiliary target, given by $\pi_i^{\mathcal{H}} \sim q_{\psi}(\pi_i^{\mathcal{H}} \mid \mathcal{H}_i, x_i)$.

We utilize the indicator function $\mathbb{I}(\cdot)$ to signify the theorem prover's verification.
Through verifying the two formal proofs $\pi_i^{\mathcal{M}}$ and $\pi_i^{\mathcal{H}}$, we establish the combined reward for the counterexample $x_i$ as $r_i := r^{\mathcal{M}}_i + r^{\mathcal{H}}_i$, where $r^{\mathcal{M}}_i := \alpha \mathbb{I}(\pi_i^{\mathcal{M}})$, $r^{\mathcal{H}}_i := (1 - \alpha) \mathbb{I}(\pi_i^{\mathcal{H}})$, and $\alpha \in [0,1]$ is a trade-off coefficient. 
Then, the supervised fine-tuning of counterexample generation is conducted on the weighted dataset $\{(\mathcal{M}_i, x_i; r_i)\}_{i=1}^N$. 
We continue to use these two rewards as the weights for the supervised fine-tuning of formal proof generation. 
The corresponding weighted dataset is derived from the formal proofs of two theorems and structured as $\{((\mathcal{M}_i, x_i), \pi_i^{\mathcal{M}}; r^{\mathcal{M}}_i)\}_{i=1}^N \cup \{((\mathcal{H}_i, x_i), \pi_i^{\mathcal{H}}; r^{\mathcal{H}}_i)\}_{i=1}^N$.

The overall framework is accordingly summarized in Algorithm~\ref{alg:framework}.

\begin{algorithm} 
\caption{The Integrated Workflow of Data Mutation and Multi-Reward Training} 
\label{alg:framework}
\begin{algorithmic}[1]
\Require A seed dataset $\{\mathcal{T}_i\}_{i=1}^N$, with each formally provable and expressed in a universal format.
\Ensure Two LLMs, $q_{\phi}$ and $q_{\psi}$, for generating counterexamples and formal proofs. 
\vspace{0.4em}
\State \emph{\# Phase 1: Dataset Mutation}
\For{$i = 1, \dots, N$} 
    \State Mutate $\mathcal{T}_i$, collecting the counterexample target $\mathcal{M}_i$ and the auxiliary target $\mathcal{H}_i$.
\EndFor
\vspace{0.4em}
\State \emph{\# Phase 2: Expert Iteration}
\For{$k = 1, \dots$}
    \State {\emph{\# Step 2.1: Large-scale Inference and Verification}}
    \For{$i = 1, \dots, N$} 
      \State Generate a counterexample $x_i$ using $q_{\phi}$: $x_i \sim q_{\phi}(x_i \mid \mathcal{M}_i)$.
      \State Generate formal proofs for the counterexample and auxiliary targets:
      \State \quad $\pi_i^{\mathcal{M}} \sim q_{\psi}(\pi_i^{\mathcal{M}} \mid \mathcal{M}_i, x_i)$ and $\pi_i^{\mathcal{H}} \sim q_{\psi}(\pi_i^{\mathcal{H}} \mid \mathcal{H}_i, x_i)$.
      \State Compute rewards based on proof correctness:
      \State \quad $r^{\mathcal{M}}_i := \alpha \cdot \mathbb{I}(\text{Verify}(\pi_i^{\mathcal{M}}))$, $r^{\mathcal{H}}_i := (1 - \alpha) \cdot \mathbb{I}(\text{Verify}(\pi_i^{\mathcal{H}}))$, and $r_i := r^{\mathcal{M}}_i + r^{\mathcal{H}}_i$.
    \EndFor
    \State {\emph{\# Step 2.2: Supervised Model Fine-tuning}}
    \State Update the model $q_{\phi}$ based on the weighted dataset $\{(\mathcal{M}_i, x_i; r_i)\}_{i=1}^N$.
    \State Update the model $q_{\psi}$ on the dataset $\{((\mathcal{M}_i, x_i), \pi_i^{\mathcal{M}}; r^{\mathcal{M}}_i)\}_{i=1}^N \cup \{((\mathcal{H}_i, x_i), \pi_i^{\mathcal{H}}; r^{\mathcal{H}}_i)\}_{i=1}^N$.
\EndFor
\end{algorithmic}
\end{algorithm}

%% file: sections/004_experiment.tex
\section{Experiments} \label{sec:experiment}

This section presents the experimental results. Specifically, our study aims to address the following research questions (RQs):
\begin{itemize}[leftmargin=1em]
\item \textbf{RQ1 (Efficacy and efficiency of data mutation):} What is the success rate and time consumption of data mutation for generating new theorems from a single seed theorem?
\item \textbf{RQ2 (Efficacy and efficiency of multi-reward training):} Compared to single-reward training, does multi-reward training improve the performance and training efficiency?
\item \textbf{RQ3 (Performance improvement of integrated workflow):} Compared with baselines, does our proposed framework achieve better performance in generating formal counterexamples?
\end{itemize}

All experiments were conducted on a Linux server featuring two AMD EPYC 7C13 64-core processors equipped with eight high-end GPUs. 
The code, together with the experimental data, is available at \url{https://figshare.com/s/02e05a2c2945ee10dcc4}.

\subsection{RQ1: Efficacy and Efficiency of Data Mutation}

Leveraging the proposed mutation strategy, we can curate a substantial number of counterexample problems as training materials for LLMs. 
To obtain the seed theorems, we aggregated all theorems and lemmas from the Lean 4 formal library Mathlib~\citep{mathlib2020community} and the popular formal dataset Leanworkbook~\citep{ying2024lean}.
We also extracted intermediate subgoals or lemmas from formal proofs generated by LLMs (Goedel prover v2~\citep{lin2025goedel} and DSP+~\citep{cao2025reviving}) on the MiniF2F~\citep{zheng2021minif2f} and PutnamBench~\citep{tsoukalas2024putnambench}. 
These four data sources originate from diverse mathematical domains and varying levels of difficulty. 
Mathlib encompasses a broad spectrum of mathematical concepts and foundational lemmas, while Leanworkbook comprises formalized content from approximately ten distinct categories (e.g., algebra, number theory, combinatorics). 
MiniF2F and PutnamBench focus on competition-level problems, with the former targeting high-school-level challenges and the latter undergraduate-level tasks, potentially involving more complex concepts.
 
We implemented a Lean 4 tactic, \emph{mutate}, to perform the symbolic mutation. 
Before applying it, we utilized the Lean 4 theorem prover to analyze the formal proof of the seed theorem, identifying and eliminating hypotheses unused in that proof.
Following the execution of the mutate operator, the theorem prover was invoked again to ensure the grammatical correctness of the newly generated problem. 

The results of the mutation process on the four data sources are presented in Table~\ref{tab:data_mutation}.
In total, we collected approximately 575K candidate counterexample problems, which constitute a sufficiently large dataset to support LLM fine-tuning.
The efficacy and efficiency of the mutation procedure are illustrated in Figure~\ref{fig:data_mutation}.
The proposed method yields a mutation ratio in the approximate range of 1.6--2.5, while the average execution time per seed theorem remains between 0.3 and 0.71 seconds.
Note that the execution time on Mathlib is comparatively lower because this formal library is pre-compiled.

\input{results/mutation.tex}

\subsection{RQ2: Efficacy and efficiency of multi-reward training}

To evaluate the proposed multi-reward training, we conducted an ablation study by comparing training with and without the multi-reward function.
To elaborate, we selected Qwen3 8B as the base model~\citep{qwen3technicalreport} for informal reasoning, which is responsible for proposing counterexamples, 
and DeepSeek-Prover-v2 7B~\citep{ren2025deepseek} as the base model for formal reasoning, which generates proofs for the proposed counterexamples.
Both models were trained under the framework described in Algorithm~\ref{alg:framework}, with the sole difference between the single-reward and multi-reward settings being the choice of coefficient $\alpha$.
In the single-reward setting, $\alpha$ was fixed at $1.0$, whereas in the multi-reward setting, it was set to $0.8$ to ensure that both rewards contribute effectively.
We also adapted Dr.GRPO~\citep{liu2025understanding}, a normalized variant of GRPO, to regulate proof length, thereby preventing proofs from being excessively short (overfitting) or unnecessarily long (overthinking).

From the collected dataset of 575K instances, we reserved 3K for validation. 
Due to limited computational resources, training was limited to 56 iterations with a batch size of 10K.
In each iteration, large-scale inference and verification were conducted, after which both models were updated accordingly.

\begin{figure}[t]
    \centering
    \includegraphics[width=0.98\textwidth]{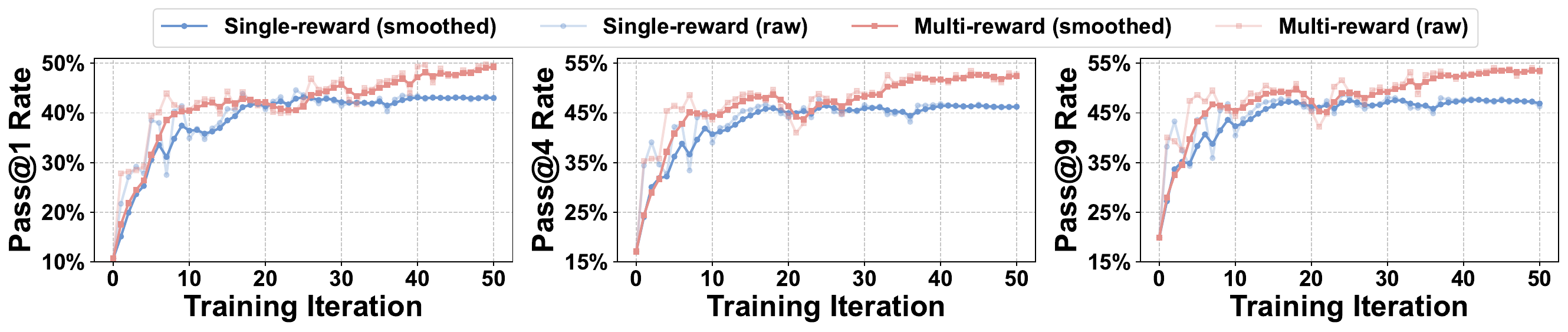}  
    \caption{Pass@\textit{k} ($k=1,4,9$) rate curves on the validation set. Compared with the single-reward baseline, multi-reward training converges faster and yields superior final performance.}

    \label{fig:curves}
\end{figure}

We compute pass@\textit{k} rates for $k=1, 4, 9$ on the validation set and plot the pass@\textit{k} rate curves in Figure~\ref{fig:curves}. 
Compared with the single-reward training, the multi-reward approach achieves both faster convergence and higher final performance.
Specifically, at convergence, multi-reward training reaches approximately 49\%, 52\%, and 54\% for pass@1, pass@4, and pass@9, respectively, whereas the single-reward baseline attains only 43\%, 46\%, and 47\%.
For reference, we also report the pass@\textit{k} results of existing LLM-based theorem provers in Appendix~\ref{app:additional_results}.

\subsection{RQ3: Performance improvement of integrated workflow}

\newcommand{\counter}{\textsc{For-Counter}\xspace}
\newcommand{\veriR}{\textsc{Veri-Reason}\xspace}
\newcommand{\veriF}{\textsc{Veri-Formalize}\xspace}

We establish three tasks to evaluate our fine-tuned model: the identification of counterexamples, as well as the verification of autoformalized results and reasoning steps.
For the counterexample identification, we employ Kimina-Autoformalizer 7B~\citep{wang2025kimina} to formalize CounterMath~\citep{li2025one}, which is constructed from four classic counterexample textbooks.
This yields 1,058 formal counterexample problems, forming our benchmark (denoted as \counter).
Furthermore, we assess the effectiveness of counterexample generation by identifying potential errors in the formalization of correct theorems and in the reasoning steps of proving theorems.
Therefore, we construct 
\veriF, which comprises 3K unprovable problems generated by formalizing FormalMath~\citep{yu2025formalmath}, 
and another 3K dataset \veriR obtained when applying DSP+~\citep{cao2025reviving} to solve the problems from FormalMath.

\input{results/comparison.tex}

The pass@\textit{k} ($k=1,4,9$) results are summarized in Table~\ref{tab:results}.
Instead of reporting pass@$k$ rates, we present the absolute number of correctly solved problems, since the established benchmark contains inaccuracies.
The results show that the model fine-tuned with our framework substantially outperforms existing state-of-the-art LLMs.
Overall, open-source neural provers demonstrate stronger performance than general-purpose LLMs.
Our fine-tuned model further advances the state of the art: for pass@1, it solves 95, 69, and 63 more problems than the strongest baseline across the three benchmarks.
Similar gains are observed for pass@4 and pass@9, further confirming the superiority of our model.
We also conduct an ablation study in Appendix~\ref{app:additional_results} to evaluate the contribution of each component (i.e., informal and formal reasoning models).

%% file: results/mutation.tex
\begin{table}[t]
\begin{minipage}[t]{0.46\textwidth}
\captionsetup{type=table}
\centering
\begin{threeparttable}
\resizebox{0.98\linewidth}{!}{
\begin{tabular}{lrr}
\toprule 
\multirow{2}*{Source} & \# of seed & \# of mutated \\
&  theorems  &  problems \\
\midrule
Mathlib & 43,990 & 88,217 \\
Leanworkbook & 17,061 & 42,480 \\
MiniF2F & 48,415 & 94,215 \\
PutnamBench & 212,463 & 350,127 \\
\midrule
Total & 321,929 & 575,039 \\
\bottomrule
\end{tabular}
}
\caption{Overall results of data mutation. Applied to a set of seed theorems, our mutation method generates a diverse collection of candidate counterexample problems.}
\label{tab:data_mutation}
\end{threeparttable}
\end{minipage}
\hfill
\begin{minipage}[t]{0.52\textwidth}
\captionsetup{type=figure}

\includegraphics[width=\linewidth]{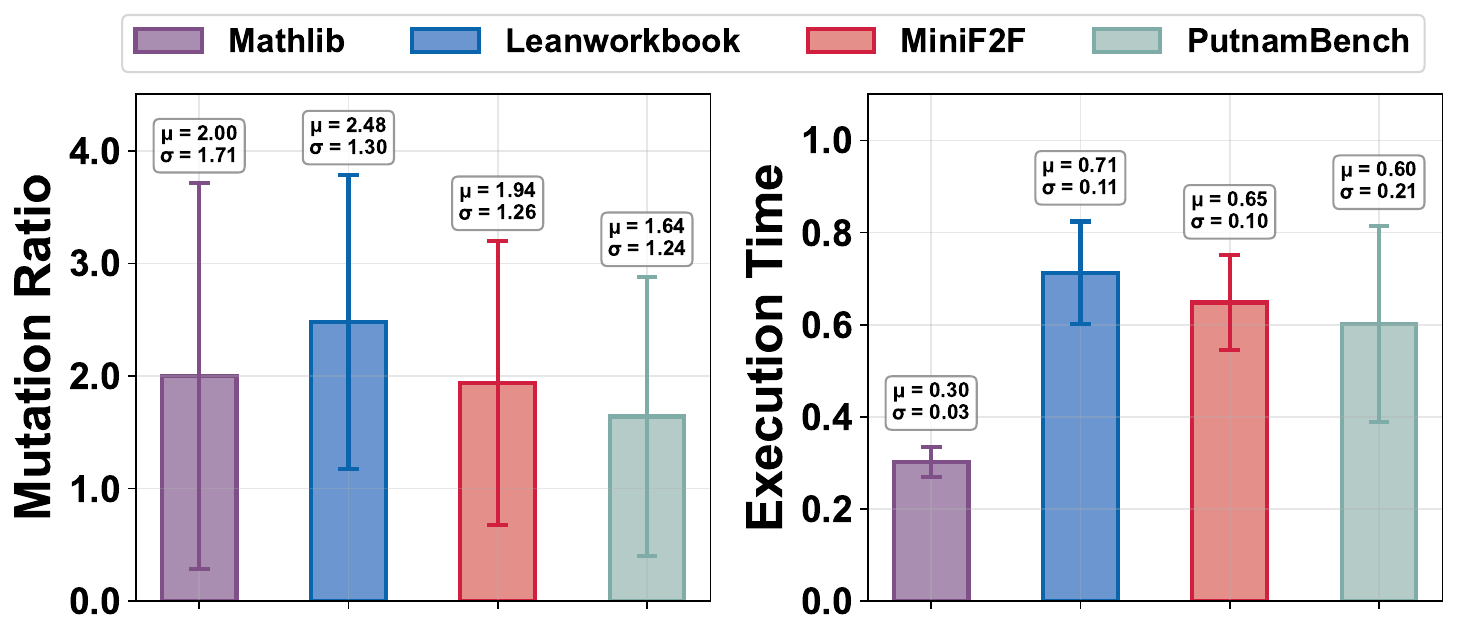}

\caption{Results of mutation ratio and average execution time. Evaluated across multiple data sources, our method maintains a high mutation ratio while remaining computationally efficient.}
\label{fig:data_mutation}
\end{minipage}
\end{table}

%% file: results/comparison.tex
\newcommand{\pass}{%
  \tikz[baseline=-0.9ex,scale=0.3] {%
    \draw[line width=1.2pt, line cap=butt, line join=miter]
      (0,0) .. controls (0.05,-0.05) and (0.18,-0.22) .. (0.35,-0.45)
             -- (0.95,0.35);
  }%
}
\newcommand{\compose}{%
  \tikz[baseline=0.1ex,scale=0.2] {%
    \path[use as bounding box] (0.4,0) rectangle (1.2,1);
    \draw[line width=0.8pt, line cap=round]
      (0.25,0.25) -- (0.75,0.75)
      (0.25,0.75) -- (0.75,0.25);
  }%
}

\begin{table}[t]
    \centering
    \caption{Numbers of solved problems on three counterexample-generation tasks under pass@\textit{k} ($k=1,4,9$). Relative percentage improvements in the $\Delta$ row are rounded down to the nearest integer. Our fine-tuned model achieves the best performance on all three benchmarks and all three \textit{k} values, outperforming both general-purpose LLMs and open-source neural theorem provers.}

    \label{tab:results}
    \begin{threeparttable}
    \resizebox{1.0\linewidth}{!}{
    \begin{tabular}{c|rrr|rrr|rrr}
    \toprule 
    \multirow{2.5}*{Dataset} & \multicolumn{3}{c|}{\counter} & \multicolumn{3}{c|}{\veriR} & \multicolumn{3}{c}{\veriF} \\
     & \pass @1 & \pass @4 & \pass @9 & \pass @1 & \pass @4 & \pass @9 & \pass @1 & \pass @4 & \pass @9 \\
    \midrule
    \multicolumn{10}{c}{\sc General-purpose LLMs} \\
    Gemini-2.5-Flash & 21 & 66 & 82 & 5 & 9 & 15 & 2 & 8 & 13 \\
    Grok-3-mini & 30 & 74 & 101 & 4 & 10 & 19 & 2 & 8 & 19 \\
    GPT-4.1-mini & 19 & 65 & 103 & 54 & 97 & 150 & 31 & 87 & 137 \\
    Deepseek-R1 & 61 & 135 & 158 & 51 & 75 & 105 & 27 & 72 & 102 \\
    \midrule
    \multicolumn{10}{c}{\sc Open-sourced Neural Provers} \\
    Leanabell-prover & 106 & 159 & 181 & 144 & 210 & 231 & 111 & 198 & 228 \\
    STP-prover & 101 & 157 & 179 & 131 & 151 & 170 & 99 & 150 & 171 \\
    Kimina-prover-distill & 31 & 109 & 165 & 66 & 114 & 156 & 18 & 141 & 249 \\
    Goedel-prover-v2 & 89 & 177 & 215 & 88 & 147 & 200 & 63 & 165 & 201 \\
    Deepseek-prover-v2 & 127 & 200 & 224 & 144 & 203 & 234 & 69 & 135 & 186 \\
    \midrule 
    Ours & \textbf{222} & \textbf{274} & \textbf{302} & \textbf{213} & \textbf{260} & \textbf{295} & \textbf{174} & \textbf{255} & \textbf{313} \\
\rowcolor[gray]{.92} 
$\Delta$ & 
\makecell[r]{95 \\ {\small 74\%$\uparrow$}} & 
\makecell[r]{74  \\ {\small 37\%$\uparrow$}}  & 
\makecell[r]{78  \\ {\small 34\%$\uparrow$}}  & 
\makecell[r]{69 \\ {\small 47\%$\uparrow$}}   & 
\makecell[r]{50 \\ {\small 23\%$\uparrow$}}   & 
\makecell[r]{61 \\ {\small 26\%$\uparrow$}} & 
\makecell[r]{63 \\ {\small 56\%$\uparrow$}}   & 
\makecell[r]{57 \\ {\small 28\%$\uparrow$}}   & 
\makecell[r]{64 \\ {\small 25\%$\uparrow$}} \\
    \bottomrule
    \end{tabular}
    }
    {\footnotesize
    }
    \end{threeparttable}
\end{table}

%% file: sections/005_relatedwork.tex
\section{Related Work}

Formal counterexample generation is generally related to LLM-based counterexample generation and formal reasoning (a.k.a neural theorem proving).
We provide a brief overview of related work in these directions and highlight the connections and differences between our work and theirs.



\subsection{LLM-based Counterexample Generation}
Counterexample finding is essential for both advancing mathematical research and strengthening the reasoning capabilities of LLMs. 
Despite its importance, there has been relatively little work on incorporating counterexample generation and AI techniques.
Early attempts~\citep{wagner2021constructions, ghebleh2024reinforcement, ghebleh2024reinforcement2, roucairol2022refutation, vito2023adaptive} employed reinforcement learning and Monte Carlo tree search to identify counterexamples in combinatorics and graph theory, later extending this line of work with transformer-based models to improve search performance~\citep{charton2024patternboost}.
More recently, some studies have begun to explore the potential of LLMs in proposing counterexamples.
For example,
CounterMath~\citep{li2025one} introduced a benchmark in this direction, demonstrating that current LLMs still face significant challenges.
However, this benchmark is established purely in natural language, making it difficult to rigorously verify the correctness of counterexample candidates.
MathConstruct~\citep{balunovic2025mathconstruct}, by contrast, evaluates constructive proof generation and integrates symbolic methods for validation. 
Compared with our work, it is confined to competition-level problems, whereas our work grounds the task in Lean 4 formal language and focuses on undergraduate-level settings.

\subsection{LLM-based Formal Reasoning}
The mathematical reasoning capability of LLMs is widely regarded as a key indicator of their overall intelligence~\citep{ahn2024large, asperti2025thinking, pantsar2025recognize}.
Formal reasoning requires an LLM to express its thought process in a formal language, with theorem provers verifying each step~\citep{yang2024formal, li2024survey}.
Leveraging advanced techniques such as autoformalization~\citep{wu2022autoformalization}, premise retrieval~\citep{yang2023leandojo}, library learning~\citep{wang2023lego}, and proof search~\citep{lample2022hypertree}, LLM-based formal reasoning has achieved remarkable progress~\citep{ji2025leanabell, cao2025reviving, ren2025deepseek, lin2025goedel}, culminating in silver-medal level performance at the 2025 International Mathematical Olympiad~\citep{chen2025seed}.

However, to the best of our knowledge, systematic investigations into the formal proof of counterexample problems remain extremely limited. 
Although some methods involve disproving the newly collected theorem when it cannot be successfully proved~\citep{jiang2024leanreasoner, lin2025goedel, ren2025deepseek}, the majority of available data do not fall into this category.
As a result, existing LLM-based theorem provers perform poorly on counterexample-guided formal reasoning.
Our work explicitly targets this gap by synthesizing counterexample problems and fine-tuning the LLMs.
Additionally, our work follows the informal-to-formal paradigm for formal proof generation.
Different from prior approaches that tightly align informal and formal reasoning~\citep{jiang2022draft}, we adopt a sequential scheme: the LLM first engages in informal reasoning to search for counterexamples, while formal reasoning is employed to verify the validity of the candidate results through proof.

\subsection{Formal Data Synthesis}
Data plays a crucial role in enhancing the reasoning capability of LLMs~\citep{zeng2024skywork, yang2024formal, fleureau2024numinamath}. 
Beyond manual collection and annotation, formal data synthesis methods are typically categorized into two categories: neural and symbolic.
Neural methods employ LLMs to either autoformalize existing informal problems or directly generate formal ones.
For instance, Minimo~\citep{poesia2024learning} and STP~\citep{dong2025stp} use an LLM conjecturer to propose new theorems, while DeepSeek-Prover~\citep{xin2024deepseek}, Kimi-Prover~\citep{wang2025kimina}, and Goedel~\citep{lin2025goedelv1} fine-tune autoformalization models to translate math competition problems into Lean theorems.
Symbolic methods often employ a precise and controllable mutation mechanism to ensure the validity of the generated problems. 
MetaGen~\citep{wang2020learning} mutates Metamath theorems via substitution, and works such as LeanNavigator~\citep{yin2025generating} and Alchemy~\citep{wu2024alchemy} extend this to Lean 4 by applying tactics like \emph{apply} or \emph{rw}.
Although broadly applicable, these operators often fail to produce sufficiently hard problems.
Thus, some focus on domain-specific strategies, e.g., geometry~\citep{trinh2024solving}, inequalities~\citep{wei2024proving}, or TCS~\citep{zhang2025lean}, yielding stronger training performance.

Our proposed method falls into the second category; we target the counterexample problem, devise a dedicated mutation strategy, and implement it as a new Lean 4 tactic.
This generates syntactically valid candidate problems and can be applied to neural-generated theorems as well; the existence of a counterexample still requires verification.
Based on the mutation strategy, we further design a multi-reward function compatible with popular reinforcement learning methods like GRPO~\citep{shao2024deepseekmath} and its variants~\citep{zhang2025survey}.

%% file: sections/00x_appendix.tex
\newtcolorbox{promptbox}[2][]{
    enhanced,breakable,lines before break=5,
    colback=black!1,colframe=black!25,colbacktitle=black!5,coltitle=black,
    fonttitle=\small\bfseries,title=#2,title after break={#2 (continued)},
    top=5pt,bottom=5pt,left=8pt,right=8pt,boxrule=0.4pt,arc=1.5pt,
    before skip=10pt,after skip=10pt,#1
}

\section{Usage of LLMs in Paper Writing}
We used GPT-4o solely for grammar checking and minor language polishing. The LLM was not involved in research ideation, data analysis, or substantive writing. The authors take full responsibility for the content of this paper.

\section{Illustrative Examples}

\begin{figure}[thb]
    \centering
    \includegraphics[width=1.0\textwidth]{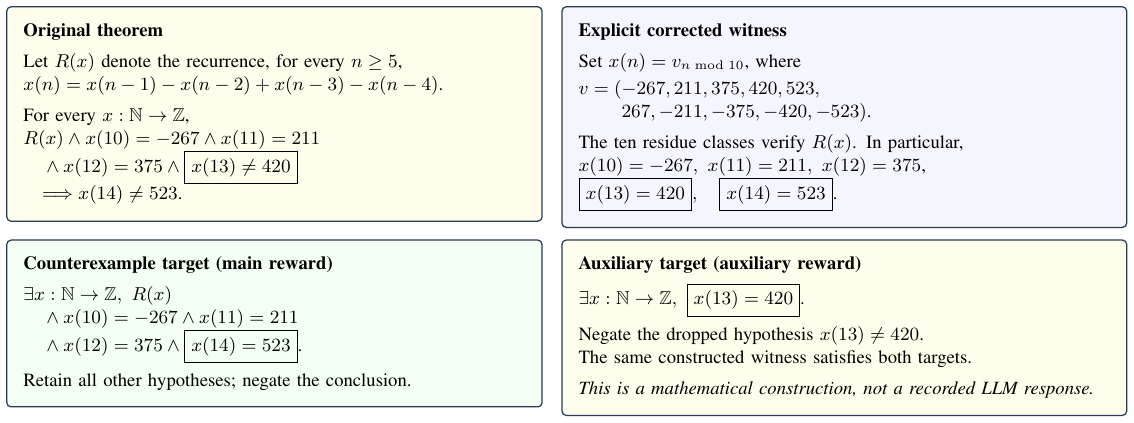}  
    \caption{
    A corrected mathematical illustration of our counterexample training framework. The seed theorem is an intermediate subgoal extracted from DSP+'s proof of AIME-II 2001 Problem 3 in miniF2F. The witness is an explicit construction, not a recorded model response.
    }
    \label{fig:detailed_framework}
\end{figure}

We further illustrate our framework using a concrete example in Figure~\ref{fig:detailed_framework}.
Starting from the original theorem \leaninline{aimeII_2001_p3_g4_extracted_54}, we discard \leaninline{h@@u2085@@ : x 13 @@u2260@@ 420}. The counterexample target \leaninline{aimeII_2001_p3_mut_54_drop4} retains the recurrence and the values of $x(10),x(11),x(12)$, but negates the conclusion to require $x(14)=523$. The auxiliary target \leaninline{aimeII_2001_p3_54_drop4} negates the dropped hypothesis to require $x(13)=420$. Figure~\ref{fig:detailed_framework} gives a single witness satisfying both targets.

\section{Experimental Setting} \label{app:setting}
In our expert iteration framework, we employ two LLMs; Qwen3 8B~\citep{qwen3technicalreport} for generating candidate counterexamples and Deepseek-Prover-V2 7B~\citep{ren2025deepseek} for producing formal proofs.
We fix the maximum generation length to 4,096 tokens and set the sampling temperature to 0.9 for both models.
For weighted training, we adapt LLaMA-Factory~\citep{zheng2024llamafactory}, using its default configuration (Adam optimizer~\citep{kingma2014adam} with a learning rate of 1e-5 and cosine learning rate schedule).
Following prior work~\citep{dong2025stp, lin2025goedel}, we perform an additional re-training step on the collected dataset paired with verified correct answers.

During evaluation, for each problem, we prompt Qwen3 to generate three counterexample candidates, and for each candidate, Deepseek-Prover-V2 produces three corresponding formal proofs.
For verification, we employ lean-repl with a 60-second timeout and an 8GB memory limit per proof.

Regarding the prompt setting, for the baseline methods, we adopt the official prompt templates provided in their respective repositories.
In our method, the prompts are designed for Qwen3 and Deepseek-Prover-V2, and are detailed in the following. 

{\small
\begin{promptbox}{Prompt of Informal Reasoning for Counterexample Search}
Find a concrete example to prove the following existential problem. \\
Note that: \\
1. Please reason the problem and give the final answer in Natural Language. \\
2. The final answer should be in the format \textbackslash \textbackslash  boxed\{\{...\}\}. \\
The problem is: \{formal\_statement\}
\end{promptbox}
}

{\small
\begin{promptbox}{Prompts of Formal Reasoning for Proof Generation}
Complete the following Lean 4 code using the given concrete example \{example\}:
\begin{lstlisting}[language=lean]
```lean4
{header}
{formal_statement}
\end{lstlisting}
\end{promptbox}
}

\section{Additional Results} \label{app:additional_results}

\begin{table}[h!]
    \centering
    \caption{Pass@\textit{k} ($k=1,4,9$) rates on the validation set. The results demonstrate that our framework enables the fine-tuned model to substantially outperform baselines.}
    \label{tab:performance_on_validation}
    \begin{threeparttable}
    \resizebox{1.0\linewidth}{!}{
    \begin{tabular}{cccccc|rr}
    \toprule 
    Model & Leanabell & STP & Kimina-distill & Goedel-v2 & Deepseek-v2 & Ours & $\Delta$ (pp) \\
    \midrule
    \pass @1  & 25.4\% & 23.7\% & 9.8\% & 14.0\% & 14.1\% & \textbf{49.8\%} & {\cellcolor[gray]{0.9} 24.4$\uparrow$} \\
    \pass @4  & 36.0\% & 33.2\% & 25.3\% & 31.1\% & 30.0\% & \textbf{52.7\%} & {\cellcolor[gray]{0.9} 16.7$\uparrow$} \\
    \pass @9  & 39.9\% & 37.5\% & 35.8\% & 38.2\% & 36.2\% & \textbf{54.1\%} & {\cellcolor[gray]{0.9} 14.2$\uparrow$} \\
    \bottomrule
    \end{tabular}
    }
    {\footnotesize
    }
    \end{threeparttable}
\end{table}

We also evaluate the performance of our fine-tuned model against existing neural theorem provers on the split-out validation set.
As shown in Table~\ref{tab:performance_on_validation}, the fine-tuned model substantially outperforms the existing approaches by a large margin (14.2 to 24.4 percentage points).
This improvement can be attributed to our synthesized dataset and the effectiveness of multi-reward training.

\begin{figure}[h!]
    \centering
    \begin{subfigure}{0.31\textwidth}
        \centering
        \includegraphics[width=\linewidth]{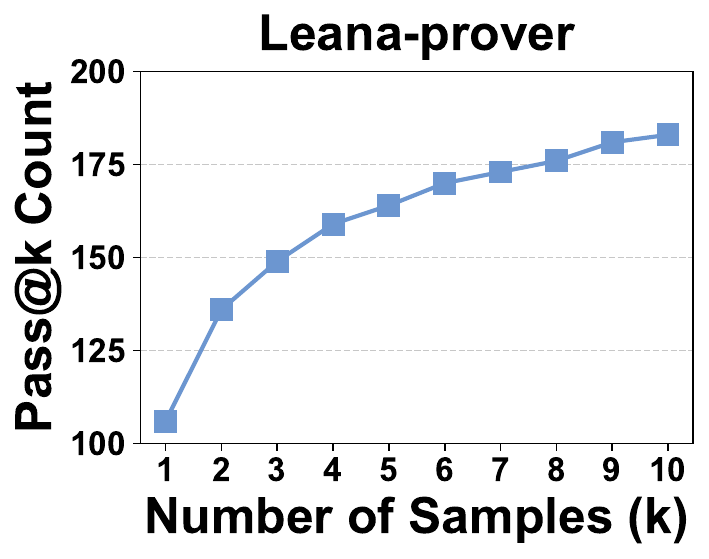}
        \label{fig:subfig-a}
    \end{subfigure}\hfill
    \begin{subfigure}{0.31\textwidth}
        \centering
        \includegraphics[width=\linewidth]{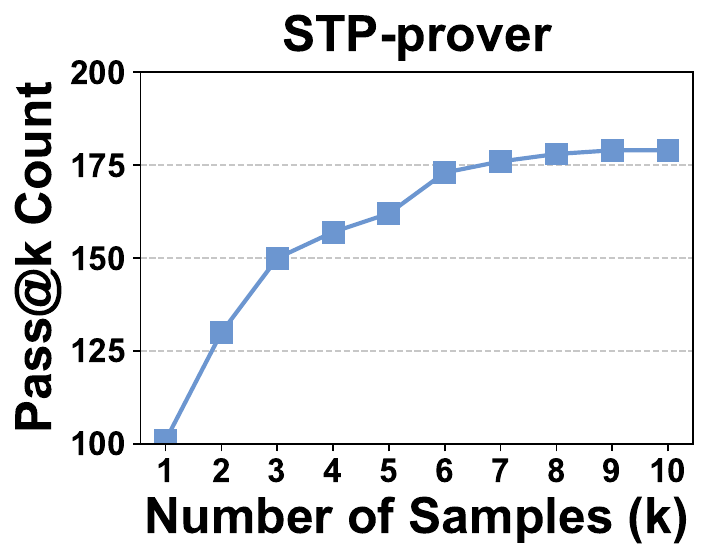}
        \label{fig:subfig-b}
    \end{subfigure}\hfill
    \begin{subfigure}{0.31\textwidth}
        \centering
        \includegraphics[width=\linewidth]{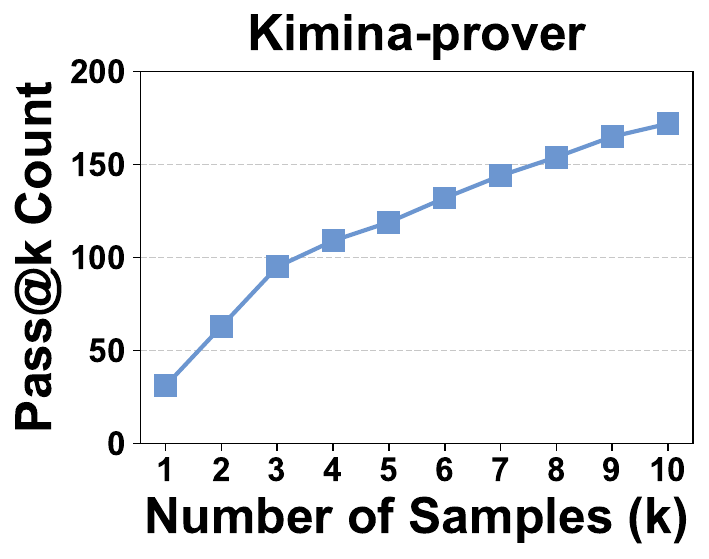}
        \label{fig:subfig-c}
    \end{subfigure}\par\medskip
    \begin{subfigure}{0.31\textwidth}
        \centering
        \includegraphics[width=\linewidth]{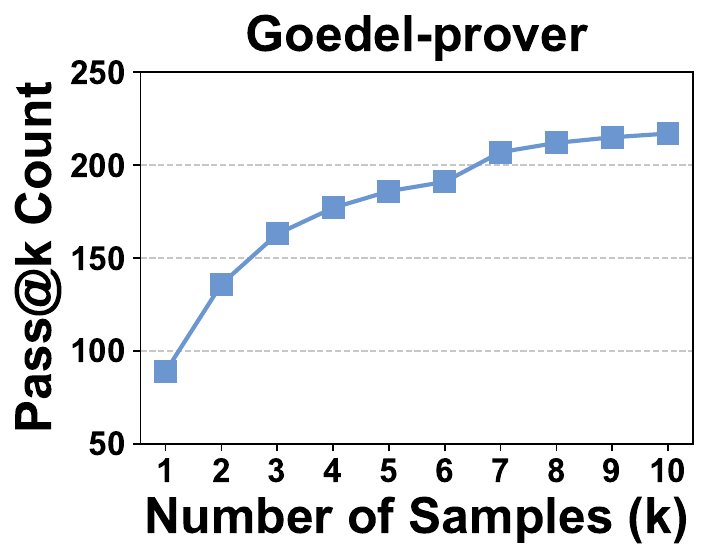}
        \label{fig:subfig-d}
    \end{subfigure}\hspace{0.025\textwidth}
    \begin{subfigure}{0.31\textwidth}
        \centering
        \includegraphics[width=\linewidth]{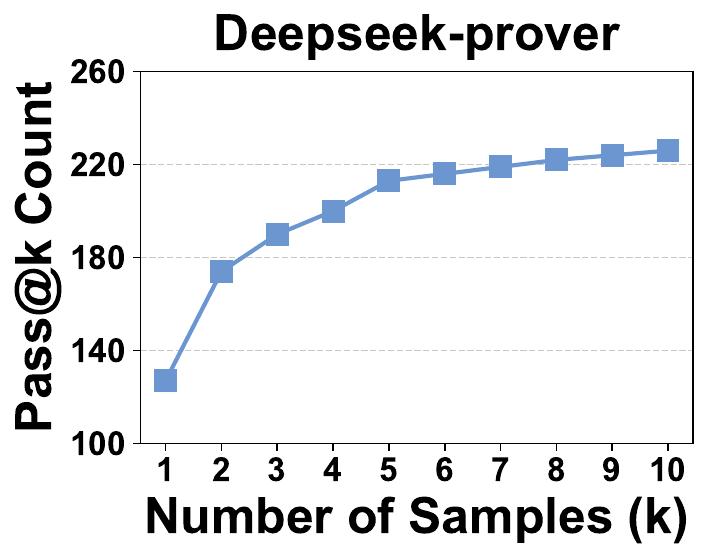}
        \label{fig:subfig-e}
    \end{subfigure}

    \caption{Pass@\textit{k} curves of five neural theorem provers. The results are derived from the evaluation of \counter and show that the performance nearly converges when $k=10$.}
    \label{fig:whole}
\end{figure}

We plot the pass@\textit{k} curves in Figure~\ref{fig:whole} to explain why we select only $k=1,4,9$ in our empirical evaluation.
The results show that the performance of the five existing neural theorem provers nearly converges once $k$ approaches 10.
Therefore, using $k=1,4,9$ is sufficient to capture the performance differences in our comparison.

\begin{table}[h!]
    \centering
    \caption{Pass@\textit{k} ($k=1,4,9$) rates for different module combinations. Results evaluated on the validation set and \counter show that both modules (informal reasoning and formal reasoning models) are successfully improved by fine-tuning with our framework.}
    \label{tab:performance_on_comb}
    \begin{threeparttable}
    \resizebox{1.0\linewidth}{!}{
    \begin{tabular}{ccccc|r}
    \toprule 
    Model & Pass@\textit{k} & Qwen3 \compose Ds-prv-v2 & Ours \compose Ds-prv-v2 & Qwen3 \compose Ours & Ours \\
    \midrule
    \multirow{3}*{Validation Set} 
    & \pass @1 & 26.2\% & 30.9\% & 47.2\% & 49.8\% \\
    & \pass @4 & 36.2\% & 38.7\% & 50.3\% & 52.7\% \\
    & \pass @9 & 40.4\% & 43.7\% & 51.0\% & 54.1\% \\
    \midrule
    \multirow{3}*{\counter} 
    & \pass @1 & 14.1\% & 15.2\% & 19.1\% & 20.9\% \\
    & \pass @4 & 18.9\% & 20.1\% & 24.3\% & 25.8\% \\
    & \pass @9 & 21.0\% & 21.5\% & 27.7\% & 28.5\% \\
    \bottomrule
    \end{tabular}
    }
    {\footnotesize
    }
    \end{threeparttable}
\end{table}

We conduct an ablation study to evaluate the contribution of each component to the overall performance. 
Specifically, our method first employs the fine-tuned Qwen3 to propose a candidate counterexample, and then leverages Deepseek-prover-v2 (denoted as Ds-prv-v2) to generate the formal proof based on this counterexample. 
To examine the necessity and effectiveness of each component, we replace the fine-tuned Qwen3 with its original version, and Ds-prv-v2 with its original version, respectively.
We also include the evaluation results for the combination of the original Qwen3 and the original Ds-prv-v2 as a reference.

The overall results are presented in Table~\ref{tab:performance_on_comb}.
They demonstrate that fine-tuning both modules contributes to performance, and combining our two fine-tuned models yields the best performance.
Compared to the gains from fine-tuning the formal reasoning model, the improvements from fine-tuning the informal reasoning model are relatively marginal.
We therefore conduct case studies in the next section to analyze the underlying reasons.

\clearpage
\section{Case Studies} \label{app:case_study}

We revisit the validation-set seed theorem used in Figure~\ref{fig:detailed_framework}. We first state the corrected counterexample target, then review the earlier draft's model-output excerpts and their limitations.
The original theorem, as well as its corresponding proof, is shown below.

\begin{tcolorbox}[breakable, enhanced jigsaw, top=5pt, bottom=5pt]
\small{
\begin{lstlisting}[style=lean, frame=none]
theorem aimeII_2001_p3_g4_extracted_54
    (x : @@u2115@@ @@u2192@@ @@u2124@@)
    (h@@u2081@@ : @@u2200@@ n @@u2265@@ 5, x n = x (n - 1) - x (n - 2) + x (n - 3) - x (n - 4))
    (h@@u2082@@ : x 10 = -267)
    (h@@u2083@@ : x 11 = 211)
    (h@@u2084@@ : x 12 = 375)
    (h@@u2085@@ : x 13 @@u2260@@ 420) :
    x 14 @@u2260@@ 523 := by
  have h6 : x 13 = x 12 - x 11 + x 10 - x 9 := by
    rw [h@@u2081@@ 13 (by omega)]
  have h7 : x 14 = x 13 - x 12 + x 11 - x 10 := by
    rw [h@@u2081@@ 14 (by omega)]
  rw [h7]
  rw [h@@u2082@@, h@@u2083@@, h@@u2084@@]
  simp
  intro h
  have : x 13 = 420 := by
    linarith
  exact h@@u2085@@ this
\end{lstlisting}
}
\end{tcolorbox}

Dropping the last hypothesis \leaninline{h@@u2085@@ : x 13 @@u2260@@ 420} and negating the remaining universal conjecture gives the following counterexample target. The original conclusion above remains unchanged.
\begin{tcolorbox}[breakable, enhanced jigsaw, top=5pt, bottom=5pt]
\small{
\begin{lstlisting}[style=lean, frame=none]
theorem aimeII_2001_p3_mut_54_drop4 : 
  @@u2203@@ (x : @@u2115@@ @@u2192@@ @@u2124@@),
    (@@u2200@@ n @@u2265@@ 5, x n = x (n - 1)  - x (n - 2) + x (n - 3)  - x (n - 4)) 
    @@u2227@@ x 10 = -267  @@u2227@@ x 11 = 211  @@u2227@@ x 12 = 375 @@u2227@@ x 14 = 523 := by
  sorry
\end{lstlisting}
}
\end{tcolorbox}

A concrete witness is the period-$10$ integer sequence $x(n)=v_{n\bmod 10}$ with
\[
v=(-267,211,375,420,523,267,-211,-375,-420,-523).
\]
Checking the ten residue classes verifies the recurrence. This gives $x(10)=-267$, $x(11)=211$, $x(12)=375$, $x(13)=420$, and $x(14)=523$, satisfying both the counterexample and auxiliary targets.

\paragraph{Earlier model-output excerpts.}
The prompts and responses below are retained from the earlier draft. Their displayed target incorrectly used $x(14)\ne 523$, so they must not be read as verified runs on the corrected target. These excerpts illustrate arithmetic and instruction-following errors; they do not establish performance on the corrected target. We preserve their formulas verbatim to distinguish recorded text from the corrected construction above.

{\small
\begin{promptbox}[breakable=false]{Earlier Informal Prompt (Incorrect Target)}
Find a concrete example to prove the following existential problem. \\
Note that: \\
1. Please reason the problem and give the final answer in Natural Language. \\
2. The final answer should be in the format \textbackslash \textbackslash  boxed\{\{...\}\}. \\
The problem is: 

\begin{lstlisting}[style=lean, frame=none]
theorem aimeII_2001_p3_mut_54_drop4 : 
  @@u2203@@ (x : @@u2115@@ @@u2192@@ @@u2124@@),
    (@@u2200@@ n @@u2265@@ 5, x n = x (n - 1)  - x (n - 2) + x (n - 3)  - x (n - 4)) 
    @@u2227@@ x 10 = -267  @@u2227@@ x 11 = 211  @@u2227@@ x 12 = 375 @@u2227@@ x 14 @@u2260@@ 523 := by 
  sorry
\end{lstlisting}
\end{promptbox}
}

\begin{promptbox}{Response of Informal Reasoning Model (Incorrect)}
{\small
We are given the recurrence relation
\[
x(n) = x(n-1) - x(n-2) + x(n-3) - x(n-4), \quad n \geq 5
\]
together with the constraints
\[
x(10) = -267, \quad x(11) = 211, \quad x(12) = 375, \quad x(14) \neq 523.
\]

\textbf{Step 1. Characteristic polynomial.}  
The recurrence has characteristic equation
\[
r^4 - r^3 + r^2 - r = r(r-1)(r^2+1) = 0,
\]
with roots $0,1,i,-i$. Thus the general solution is
\[
x(n) = A + C i^n + D(-i)^n.
\]

\textbf{Step 2. Apply conditions.}  
For $n=10,11,12$ we compute:
\[
\begin{aligned}
x(10) &= A - C + D = -267, \\
x(11) &= A - iC - D = 211, \\
x(12) &= A + C + D = 375.
\end{aligned}
\]

From the first and third equations: $2C=642 \implies C=321$, and $A+D=54$.  
Substituting into the second gives
\[
D = -\tfrac{157+321i}{2}, \qquad 
A = 54 - D = 54 + \tfrac{157+321i}{2}.
\]

\textbf{Step 3. Check $x(14)$.}  
Since $i^{14}=-1$ and $(-i)^{14}=-1$,
\[
x(14) = A - C - D = (A+D) - C - 2D = -267 - 2D.
\]
Plugging in $D$ yields
\[
x(14) = -110 + 321i,
\]
which is not a real number and in particular $\neq 523$.

\textbf{Conclusion.}  
A valid function $x$ exists. One explicit example is
\[
x(n) = \left(54 + \tfrac{157+321i}{2}\right) + 321 \, i^n 
       - \tfrac{157+321i}{2} \, (-i)^n.
\]
This satisfies the recurrence, the given values, and $x(14)\neq 523$.
}
\end{promptbox}

This response is incorrect: the required sequence is integer-valued, and the characteristic polynomial of the stated recurrence is $r^4-r^3+r^2-r+1$, not the polynomial in the excerpt. The next response satisfies the earlier, incorrectly stated target, but it is not a counterexample to the conjecture we intend to refute.

\begin{promptbox}{Earlier Informal Response (Previously Mislabeled Correct)}
{\small
We can explicitly construct such a sequence by fixing the first four values and then extending by the recurrence.  
Define
\[
x(1)=211,\quad x(2)=375,\quad x(3)=-103,\quad x(4)=0,
\]
and for all $n\geq 5$ set
\[
x(n) = x(n-1) - x(n-2) + x(n-3) - x(n-4).
\]

\textbf{Step 1. Compute terms.}  
Applying the recurrence successively:
\[
\begin{aligned}
& x(5) = -211+375-(-103)+0 = 267,  x(6) = -211, x(7) = -375, \\
& x(8) = 103, x(9) = 0, x(10) = 211 - 375 + (-103) - 0 = -267, \\
& x(11) = 211, x(12) = 375, x(13) = -103, x(14) = 0.
\end{aligned}
\]

\textbf{Step 2. Verify conditions.}  
\[
x(10)=-267, \quad x(11)=211, \quad x(12)=375, \quad x(14)=0\neq 523.
\]

All required conditions are satisfied.

\textbf{Remark.}  
This sequence is in fact periodic with period $10$.  
In Lean, one can constructively define $x:\mathbb{N}\to\mathbb{Z}$ either by recursion from the initial values, or directly by cases depending on $n \bmod 10$.
}
\end{promptbox}

This sequence satisfies the recurrence and the retained values, but $x(14)=0$ does not satisfy the corrected target $x(14)=523$. Its $x(13)=-103$ also fails the auxiliary target $x(13)=420$. Thus, although it solves the earlier displayed problem, it cannot be used as a counterexample to the intended conjecture. The following earlier formal response additionally changes the proposed sequence and fails the recurrence (for example at $n=10$). It is an erroneous attempt, not a verified proof.

{\small
\begin{promptbox}[breakable=false]{Earlier Formal Prompt and Response (Incorrect Target and Proof)}
Complete the following Lean 4 code using the given concrete example \\
$x(10)=-267, \quad x(11)=211, \quad x(12)=375, \quad x(14)=0\neq 523$:
\begin{lstlisting}[language=lean]
```lean4
theorem aimeII_2001_p3_mut_54_drop4 : 
  @@u2203@@ (x : @@u2115@@ @@u2192@@ @@u2124@@),
    (@@u2200@@ n @@u2265@@ 5, x n = x (n - 1)  - x (n - 2) + x (n - 3)  - x (n - 4)) 
    @@u2227@@ x 10 = -267  @@u2227@@ x 11 = 211  @@u2227@@ x 12 = 375 @@u2227@@ x 14 @@u2260@@ 523 := by 
  /-
  We need to find a function \( x : \mathbb{N} \to \mathbb{Z} \) that satisfies the recurrence relation \( x(n) = x(n - 1) - x(n - 2) + x(n - 3) - x(n - 4) \) for \( n \geq 5 \), and also satisfies \( x(10) = -267 \), \( x(11) = 211 \), \( x(12) = 375 \), and \( x(14) \neq 523 \).
  
  We can define such a function by specifying its values at each integer up to \( n = 14 \), ensuring that the recurrence relation is satisfied for \( n \geq 5 \). We then verify that the specific values for \( x(10) \) to \( x(14) \) meet the given conditions.
  -/
  use fun n => if n = 10 then -267 else if n = 11 then 211 else if n = 12 then 375 else if n = 13 then 523 else 0
  constructor
  @@u00b7@@ intro n hn
    simp [Nat.le_of_lt_succ, Nat.le_of_lt_succ] at hn
    split_ifs <;> omega
  constructor
  @@u00b7@@ simp
  constructor
  @@u00b7@@ simp
  constructor
  @@u00b7@@ simp
  simp
  <;> norm_num
  <;> omega
```
\end{lstlisting}
\end{promptbox}
}